\documentclass[10pt, conference, letterpaper]{IEEEtran}

\usepackage{graphicx}
\usepackage[misc,geometry]{ifsym} 
\usepackage{amsmath}
\usepackage{color}
\usepackage{hyperref} 
\usepackage[bottom]{footmisc}
\usepackage{caption}
\usepackage{subcaption}
\usepackage{supertabular}
\usepackage{afterpage}
\usepackage{url}
\usepackage{multicol}
\usepackage{multirow}
\usepackage{lscape}
\usepackage{float}
\usepackage{placeins}
\usepackage{afterpage}
\newcommand{\longText}[1]{\begin{tabular}[c]{@{}c@{}}#1\end{tabular}}

\usepackage{afterpage}
\usepackage{xspace}
\newcommand{\mhframework}{MH-FSF\xspace}

\title{MH-FSF: A Unified Framework for Overcoming Benchmarking and Reproducibility Limitations in Feature Selection Evaluation}

\author{
  \IEEEauthorblockN{
    \begin{minipage}[t]{0.32\linewidth}
      \centering
      Vanderson Rocha\textsuperscript{1}\\
      \footnotesize\textsuperscript{1}Federal University of Amazonas (UFAM)
    \end{minipage}
    \hfill
    \begin{minipage}[t]{0.32\linewidth}
      \centering
      Diego Kreutz\textsuperscript{2}\\
      \footnotesize\textsuperscript{2}Federal University of Pampa (UNIPAMPA)
    \end{minipage}
    \hfill
    \begin{minipage}[t]{0.32\linewidth}
      \centering
      Gabriel Canto\textsuperscript{1}\\
      \footnotesize\textsuperscript{1}Federal University of Amazonas (UFAM)
    \end{minipage}
  }
  
  \vspace{2mm} 
  
  \IEEEauthorblockN{
    \hfill
    \begin{minipage}[t]{0.32\linewidth}
      \centering
      Hendrio Bragança\textsuperscript{1}\\
      \footnotesize\textsuperscript{1}Federal University of Amazonas (UFAM)
    \end{minipage}
    \hfill
    \begin{minipage}[t]{0.32\linewidth}
      \centering
      Eduardo Feitosa\textsuperscript{1}\\
      \footnotesize\textsuperscript{1}Federal University of Amazonas (UFAM)
    \end{minipage}
    \begin{minipage}[t]{0.32\linewidth}
      \centering
    \end{minipage}
  }
}

\begin{document}

\maketitle

\begin{abstract}
Feature selection is vital for building effective predictive models, as it reduces dimensionality and emphasizes key features. However, current research often suffers from limited benchmarking and reliance on proprietary datasets. This severely hinders reproducibility and can negatively impact overall performance.
To address these limitations, we introduce the MH-FSF framework, a comprehensive, modular, and extensible platform designed to facilitate the reproduction and implementation of feature selection methods. Developed through collaborative research, MH-FSF provides implementations of 17 methods (11 classical, 6 domain-specific) and enables systematic evaluation on 10 publicly available Android malware datasets. Our results reveal performance variations across both balanced and imbalanced datasets, highlighting the critical need for data preprocessing and selection criteria that account for these asymmetries.
We demonstrate the importance of a unified platform for comparing diverse feature selection techniques, fostering methodological consistency and rigor. By providing this framework, we aim to significantly broaden the existing literature and pave the way for new research directions in feature selection, particularly within the context of Android malware detection.
\end{abstract}

\begin{IEEEkeywords}
Feature Selection, Android Malware Detection, Benchmarking, Reproducibility, Evaluation, Data Preprocessing, Machine Learning, Artificial Inteligence
\end{IEEEkeywords}

\section{Introduction}

Feature selection is crucial for constructing effective predictive models. By identifying and focusing on the most relevant feature subsets, it reduces data dimensionality, leading to improved model accuracy and significantly decreased computational overhead during training \cite{dhal2022comprehensive}. This process simplifies models, enhancing both interpretability and efficiency \cite{naheed2020importance}.

\textcolor{black}{Many feature selection studies face two persistent challenges that significantly hinder progress in the field. First, there is a heavy reliance on proprietary or inaccessible datasets, which impedes replication of experiments and/or validation of results. Second, comparative evaluations are often limited to a narrow set of baseline methods, failing to provide a comprehensive understanding of the performance of new approaches relative to the broader spectrum of available techniques. These two issues compromise the reproducibility of results and can lead to biased conclusions, limiting their adoption..}

It is crucial to acknowledge that numerous studies (e.g., \cite{mahindru2024permdroid}, \cite{mahindru2021semidroid}, \cite{wang2021new}, \cite{galib2020significant}, \cite{alazab2020rfg}, \cite{cai2021jowmdroid}, \cite{bhat2022mult}, \cite{sun2016sigpid}, \cite{wu2023droidrl}, \cite{smmarwar2022hybrid}, \cite{chimeleze2022bfedroid}, \cite{csahin2023novel}) typically compare their proposed feature selection methods with only a limited set of similar techniques. This practice, akin to the issue of dataset limitations, leads to a potentially skewed evaluation of each method's performance and benefits. Such a narrow comparison scope hinders our ability to conduct a truly fair and comprehensive assessment of feature selection methods, ultimately undermining confidence in the reported conclusions.

A persistent challenge in our field is the limited access to representative datasets for the evaluation and comparison of feature selection methods \cite{braganca2023capturing, soares2021deteccao, miranda2022debiasing, waheed2024effective, albahar2022modified, kumar2022androobfs, wang2022malradar}. The absence of standardized, publicly accessible, and current datasets hinders the consistent validation of new methodologies and compromises the reproducibility of results across various studies. Given that the diversity and representativeness of datasets are crucial for developing robust and generalizable feature selection techniques, this issue remains a substantial impediment to progress in the field.

In Android malware detection, feature selection is particularly vital due to the inherent complexity of applications. Android apps often utilize a vast array of permissions, components, and API calls, making the identification of relevant features a formidable challenge.

Our framework, MH-FSF, directly addresses the identified shortcomings in the literature by providing a structured, configurable, and extensible platform for feature selection methods. This solution fosters innovation and rigorous evaluation of new techniques, promoting the adoption of more effective and efficient methods across diverse predictive applications.

By employing MH-FSF to analyze 17 feature selection methods across 10 diverse datasets, we demonstrate that performance variations across imbalanced datasets underscore the complexity of real-world data analysis. This highlights the necessity for appropriate data preprocessing strategies or feature selection methods that account for class asymmetry, ultimately improving prediction accuracy and reliability.

Our findings reinforce the importance of platforms like MH-FSF, which integrate a wide range of selection methods, enabling direct and comprehensive comparisons. All evaluations are conducted on 10 publicly available and representative datasets, ensuring replicability and validation. This promotes standardization in feature selection research and provides a scalable, accessible resource for the community.

The main contributions of this work are twofold:
\begin{itemize}
    \item A comprehensive framework, MH-FSF, enabling the reproduction and detailed implementation of 17 feature selection methods (11 classical and 6 domain-specific).
    \item An in-depth comparative analysis of both classical and domain-specific methods, applied to 10 widely used Android malware detection datasets.
\end{itemize}

The remainder of this paper is organized as follows.
In Section \ref{sec_related_works}, we present a comprehensive review of related work, highlighting the limitations and challenges of existing feature selection methods within the context of Android malware detection.
Section \ref{sec_fs_methods} details the feature selection methods evaluated, distinguishing between classical approaches and domain-specific strategies designed for Android security.
In Section \ref{sec_tool}, we introduce the MH-FSF framework, detailing its modular architecture and functionalities.
Section \ref{sec_experimentation} describes the experimental setup, including the datasets, evaluation metrics, and machine learning models employed.
In Section \ref{sec_results}, we present and discuss the experimental results, comparing the performance of different methods across various scenarios and analyzing the impact of feature selection on malware detection.
Finally, in Section \ref{sec_final}, we provide concluding remarks, summarizing the study's contributions, the implications of our findings, and potential directions for future research.

This paper extends our earlier work presented at the Brazilian Symposium on Information Security and Computer Systems (SBSeg 2024)\footnote{\url{https://sol.sbc.org.br/index.php/sbseg_estendido/article/view/30125}}. The original paper introduced MH-FSF and summarized its core concepts; this expanded version offers a more comprehensive literature review (Sections \ref{sec_related_works} and \ref{sec_fs_methods}) and a more thorough evaluation of experimental results, including an analysis of balanced datasets in Section \ref{sec_results}. Consequently, this paper provides a broader and more nuanced understanding of the effects of feature selection strategies in Android malware detection.

\section{Related Work}
\label{sec_related_works}

Table \ref{tab_trabalhos_relacionados} summarizes related work in the field. In particular, many studies comparing feature selection methods focus predominantly on traditional techniques. The most frequently used methods include \cite{csahin2023adaption, csahin2023novel, islam2023android, mahindru2019deepdroid, zhao2015fest}:
\begin{itemize}
    \item Information Gain (IG),
    \item Chi-Square,
    \item Principal Component Analysis (PCA),
    \item Term Frequency-Inverse Document Frequency (TF-IDF), and
    \item Recursive Feature Elimination (RFE).
\end{itemize}

\begin{table*}
\small
\caption{Feature selection methods used in context of Android malware.}
\centering
\begin{tabular*}{0.82\textwidth}{c|c|c|c|c}
\multicolumn{1}{c|}{\multirow{2}{*}{\textbf{Reference}}} & \multicolumn{2}{c|}{\textbf{Methods}} &  \multicolumn{1}{c|}{\multirow{2}{*}{\textbf{Metrics}}} & \multicolumn{1}{c}{\multirow{2}{*}{\textbf{Datasets}}}
\\
\cline{2-3}
& \textbf{\#} & \textbf{Names} & & 
\\ \hline \hline
\cite{csahin2023adaption} & 8 & \longText{Info Gain (IG), Odds Ratio (OR),\\Inverse Document Frequency (IDF),\\Document Frequency Limit,\\Chi-Square, M2, ACC, ACC2} & \longText{Precision,\\Recall,\\F1 Score} &  \longText{Own\\ (APKPure,\\VirusShare)} \\ \hline
\cite{csahin2023novel} & 11 & \longText{Permissions Comparison,\\Mutual Info (MI),\\Subset CFs, Info Gain (IG),\\Random Decision Tree,\\Genetic Algorithm,\\ FF-FA-Based TF-IDF,\\IG and TF,\\Rough Set Theory (RST) \& PSO, \\ Fast Correlation\\Based Filter (FCBF),\\Linear Regression (LR)} & \longText{Precision,\\Recall,\\F1 Score} &  \longText{Own\\(APKPure)} \\ \hline
\cite{islam2023android} & 2 & RFE, PCA & \longText{Accuracy,\\Precision, Recall,\\F1 Score, R2} & \longText{CICAndMal2020,\\Drebin,\\CICMaldroid2020} \\ \hline
\cite{salah2020lightweight} & 2 & \longText{TF-IDF,\\FF-AF} & \longText{Accuracy, Precision,\\Recall, F1 Score} & Drebin \\ \hline
\cite{mahindru2019deepdroid} & 8 &  \longText{Mutual Info (MI),\\Deep Neural Network (DNN),\\Chi-Square, Gain Rate,\\ OneR, Linear Regression (LR),\\PCA, Info Gain (IG)} & \longText{Accuracy,\\F1 Score} &  \longText{Own\\(Google Play)} \\ \hline
\cite{fatima2019android} & 2 & Genetic Algorithm & \longText{ROC, Accuracy,\\Sensitivity,\\Specificity,\\Time} & \longText{Own\\(IIT Kanpur)} \\ \hline
\cite{zhao2015fest} & 3 & \longText{Chi-Square, Info Gain (IG),\\FrequenSel} & \longText{Accuracy, Precision,\\FP Rate, Recall} & Drebin \\ \hline
\cite{alomari2023malware} & 2 & \longText{LSTM,\\Multiple Correlations} & \longText{Accuracy, Precision,\\Recall, F1 Score} &  Kaggle  \\ \hline
\textbf{Our Work} & 17 & \longText{Artificial Bee Colony (ABC),\\
ANOVA, Chi-Square, LASSO,\\Info Gain (IG), PCA,\\Linear Regression (LR),\\
Mean Absolute Deviation (MAD),\\
Pearson Correlation Coef. (PCC),\\
ReliefF, RFE,\\
JOWMDroid, Multi-Tiered (MT), \\
RFG, SemiDroid,\\SigAPI, SigPID} & \longText{Accuracy, \\F1 Score, RoC,\\MCC} & \longText{Adroit, AndroCrawl,\\Android Permissions,\\DefenseDroid PI,\\DefenseDroid A\\(C, D, K),\\Drebin-215,\\KronoDroid R,\\KronoDroid E} \\
\end{tabular*}
\label{tab_trabalhos_relacionados}
\end{table*}

Among recent comprehensive studies, \cite{csahin2023novel} stands out for the evaluation of 11 feature selection methods on a proprietary dataset (APKPure) using precision, recall, and F1 score metrics. However, despite its broader scope in malware detection, several limitations impact the effectiveness and generalizability of its findings.

The first limitation lies in the dataset. 
Although it may offer valuable information about malware, the reliance on a single data source restricts the generalizability of the study. 
Furthermore, the use of a proprietary dataset, which is unavailable and lacks detailed documentation, hinders reproducibility and independent verification of results. 
There is also a risk of overfitting specific features of APKPure, which were used in the dataset's construction.

A second significant limitation is that the authors compared their method with ten other methods, each implemented and evaluated in different works using distinct datasets.
Although all studies focused on permission-type features, the lack of a common dataset undermines the comparability and reliability of the results.

With the exception of \cite{islam2023android}, the studies listed in Table \ref{tab_trabalhos_relacionados}, including \cite{salah2020lightweight}, \cite{fatima2019android}, \cite{zhao2015fest}, \cite{alomari2023malware}, and \cite{mahindru2019deepdroid}, likely share the limitations discussed previously. These works lack sufficient evidence to demonstrate the robustness of their methods across diverse datasets and mitigate potential risks. Given that dataset diversity, which includes various types and characteristics of malware, is crucial for validating the efficiency and applicability of feature selection methods in real-world scenarios, this remains a significant concern.

Our research distinguishes itself through a more comprehensive and rigorous approach compared to related works. We conduct an extensive evaluation of 17 feature selection methods, encompassing 11 classical and 6 advanced techniques, across 10 diverse datasets. Utilizing appropriate evaluation metrics, we effectively assess the generalization capabilities of these methods. This approach demonstrably enhances the robustness, reliability, and generalization of our findings, surpassing the scope and depth of existing studies.

It is important to highlight that our methodology encompasses a broader range of feature selection techniques compared to existing studies, spanning from classical methods like Information Gain (IG) and Principal Component Analysis (PCA) to advanced algorithms such as Artificial Bee Colony (ABC).

Furthermore, a key distinction of our work is the systematic reproduction, implementation, and evaluation of six sophisticated, domain-specific methods, including SemiDroid, RFG, JOWNDroid, MT, SigPID, and SigAPI.

Consequently, this study provides the first comprehensive comparative analysis of both classical and advanced feature selection methods specifically tailored for the Android malware domain.

\textcolor{black}{In the context of security, feature selection plays a strategic role, enabling the construction of leaner and more interpretable models, which are essential in decision-making processes. In addition to promoting greater transparency in detection criteria, feature selection contributes to computational efficiency, a determining factor in real-time detection systems, where response speed can directly impact attack prevention \cite{subbiah2021opportunities, thakkar2022survey}.}

\section{Feature Selection Methods}
\label{sec_fs_methods}

Feature selection methods aim to identify and select a subset of relevant features to construct predictive models. These techniques enhance model performance by eliminating irrelevant or redundant features, reducing overfitting, improving generalization, and decreasing computational overhead. In our research, we categorize these methods into two groups: classical and domain-specific (or advanced).

Classical feature selection methods have been extensively studied and applied across various domains, including malware detection. These methods encompass well-established techniques such as Principal Component Analysis (PCA), Linear Regression (LR), and Information Gain (IG).

Domain-specific feature selection methods are designed to address the unique characteristics of Android applications and the behavioral patterns of malicious apps. These methods consider both static features, such as permissions and API calls, and dynamic features, such as system call patterns. It is hypothesized that incorporating these domain-specific features enhances Android malware detection capabilities. However, the direct applicability of classical methods to Android malware detection may be limited due to the inherent complexity and domain-specific nature of Android application features.

To systematically evaluate both classical and domain-specific (advanced) feature selection methods for Android malware detection, we developed a software framework that integrates the reproduction and implementation of six advanced methods and eleven classical methods.

This collaborative effort, spanning approximately three years and involving over seven researchers, has resulted in three publications. These works detail the framework's architecture, the systematic approach to reproducing and implementing the methods reliably, and the initial validations of the early methods \cite{soares2022analise, costa2022fs3e, neves2023avaliacao}.

However, previous publications focused primarily on advanced methods within the framework, using three to five datasets. These studies provided foundational insights that informed the current work, particularly highlighting the challenge of generalizing methods across datasets from different domains.

In this study, we extended our framework by adding eleven classical feature selection methods, adhering to the rigorous procedures established in previous work, as detailed in \cite{costa2022fs3e}. For instance, each method's reproduction and implementation undergo review and technical evaluation by at least three researchers. Furthermore, we utilize ten diverse datasets, presenting a significant challenge for assessing the generalization capabilities of these methods.

Table \ref{tab_selection_methods} provides an overview of the seventeen methods now available in our software framework, accessible through a public GitHub repository \cite{rocha2024mhfsf}. 

\setlength{\tabcolsep}{5pt}
\begin{table}[ht!]
\caption{Feature selection methods implemented in MH-FSF, categorized as classical or domain-specific, including selection strategy and original reference.}
\centering
\begin{tabular}{c||c|c|c}
\textbf{Type} & \textbf{Method} & \textbf{Selection} & \textbf{Reference}\\
\hline \hline
\multirow{11}{*}{\rotatebox{90}{CLASSICS}} & Artificial Bee Colony (ABC) & Subset & \cite{karaboga2014comprehensive} \\
\cline{2-4}
 & ANOVA & Subset & \cite{st1989analysis}\\
\cline{2-4}
 & Chi-Square & Ordering & \cite{tallarida1987chi}\\
\cline{2-4}
 & Info Gain (IG) & Ordering & \cite{azhagusundari2013feature}\\
 \cline{2-4}
 & LASSO & Subset & \cite{ranstam2018lasso}\\
\cline{2-4}
 & Linear Regression (LR) & Subset & \cite{csahin2023novel} \\
 \cline{2-4}
 & Mean Absolute Deviation (MAD) & Ordering & \cite{konno2005mean}\\
\cline{2-4}
 & PCA & Subset & \cite{kurita2019principal}\\
 \cline{2-4}
 & Pearson Correlation Coef. (PCC) & Ordering & \cite{cohen2009pearson}\\
 \cline{2-4}
 & ReliefF & Ordering & \cite{robnik2003theoretical}\\
\cline{2-4}
 & RFE & Ordering & \cite{darst2018using}\\
\hline\hline
\multirow{6}{*}{\rotatebox{90}{SPECIFICS}} & JOWMDroid &Subset & \cite{cai2021jowmdroid}\\
\cline{2-4}
 & Multi-Tiered (MT) & Subset & \cite{bhat2022mult}\\
\cline{2-4}
 & RFG & Subset & \cite{alazab2020rfg}\\
\cline{2-4}
 & SemiDroid & Subset & \cite{mahindru2021semidroid}\\
\cline{2-4}
 & SigAPI & Subset & \cite{galib2020significant}\\
\cline{2-4}
 & SigPID & Subset & \cite{sun2016sigpid}\\
\end{tabular}
\label{tab_selection_methods}
\end{table}

This repository represents the third generation of our framework, incorporating substantial structural changes and the addition of eleven new methods compared to the previous version. Notably, beyond the six advanced methods listed in Table \ref{tab_selection_methods}, the framework includes additional methods such as FSDroid, ABC, and LR. These methods were excluded from the current analysis due to their consistently inferior performance compared to SemiDroid, SigPID, and RFC across at least three of the datasets used, as demonstrated in our earlier work \cite{neves2023avaliacao}.

It is noteworthy to observe in Table \ref{tab_fs_datasets} the evaluation context of the six domain-specific feature selection methods employed in this study. Originally, each of these methods was designed and evaluated using a single, proprietary dataset, often constructed by combining samples from various other datasets and sources. Datasets such as Drebin, Android Permissions Dataset (Android P. D.), and Android Malware Dataset (Android M. D.) are frequently cited as base datasets in the construction of these proprietary datasets.

In addition to the reliance on single, non-reproducible datasets, most authors utilize only imbalanced datasets. This practice can adversely affect the performance of these methods when applied to different datasets.

\setlength{\tabcolsep}{5pt}
\begin{table*}[ht!]
\caption{Datasets of methods specific to Android malware domain.}
\centering
\begin{tabular*}{.8\textwidth}{c||c|c|c|c|c}
\textbf{Method} & \textbf{Features} & \textbf{Balanced} & \textbf{\# Features} & \textbf{\# Samples} & \textbf{Dataset} \\
\hline \hline
SemiDroid & P + A & No & 1842 & 500K & Own (APKs + Android P. D.) \\ \hline
RFG & A & No & 27253 & 36915 & Own (VirusTotal + AndroZoo + etc.) \\ \hline
JOWNDroid & P + A & No & 643 & 163556 & Own (Drebin + AMD + APKs) \\ \hline
MT & P + A + I & No & 61 & 11449 & Own (Virustotal, VirusShare + Drebin) \\ \hline
SigPID & P & Yes & 135 & 310926 & Own (APKs + Android M. D.) \\ \hline
SigAPI & A & No & 142 & 18769 & Own (Drebin + Android P. D.) \\
\end{tabular*}
\caption*{[P] Permissions, [A] API Calls, [I] Intents}
\label{tab_fs_datasets}
\end{table*}

\color{black}
\section{The MH-FSF Framework}
\label{sec_tool}

Figure \ref{fig_tool_overview} illustrates the MH-FSF framework's architecture, which comprises four primary stages:

\begin{figure*}[ht!]
\centering
\includegraphics[width=1.\textwidth]{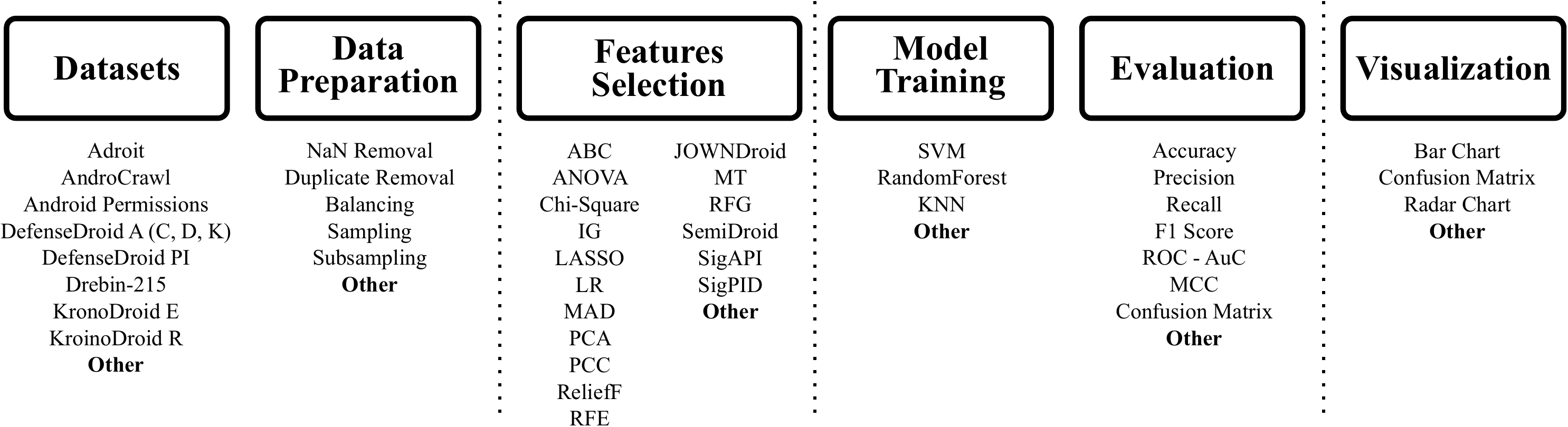}
\caption{Overview of \mhframework: The four main steps of the pipeline.}
\label{fig_tool_overview}
\end{figure*}

\begin{enumerate}
    \item Data Manipulation,
    \item Feature Selection Methods,
    \item Machine Learning Model Training and Evaluation, and
    \item Results Visualization.
\end{enumerate}

The first stage involves dataset selection and preparation, ensuring the representativeness and robustness necessary for feature selection methods. Data preparation encompasses data quality and integrity measures, including the removal of null values (NaN), duplicates, class balancing, sampling, and subsampling.

Subsequently, feature selection reduces data dimensionality and identifies the most relevant features for classification models. The MH-FSF framework currently integrates domain-specific feature selection methods tailored for Android malware detection, such as SemiDroid \cite{mahindru2021semidroid}, RFG \cite{alazab2020rfg}, JOWNDroid \cite{cai2021jowmdroid}, MT \cite{bhat2022mult}, SigPID \cite{sun2016sigpid}, and SigAPI \cite{galib2020significant}, alongside classical methods like ANOVA, Chi-Square, LASSO, PCA, ReliefF, and RFE. Each method offers a unique approach to evaluating and selecting significant features, contributing to the development of more accurate and efficient models.

Following feature selection, MH-FSF trains models using machine learning algorithms, including SVM, RandomForest, and KNN. Model evaluation is then conducted using performance metrics such as accuracy, precision, recall, F1 score, ROC-AUC, and MCC.

Finally, results are visualized through bar charts, confusion matrices, radar charts, and other visualization techniques, facilitating result interpretation, pattern identification, and clear communication of findings.

Beyond these core functionalities, MH-FSF offers extensibility, enabling the rapid and straightforward integration of new feature selection methods. This adaptability ensures the framework evolves with emerging research needs and the development of novel techniques, maintaining its relevance in the rapidly advancing field of machine learning. The ability to incorporate new methods without requiring significant structural modifications is a key factor in the framework's long-term viability and sustained relevance.

To add a new feature selection method to the framework of \mhframework, simply follow these steps:
\begin{enumerate}
    \item Create a new directory with the method identifier in the \texttt{methods} folder.
    \item Add a \texttt{about.desc} file containing the method description.
    \item Include the method code in the \texttt{run.py} file. In \texttt{run.py}, import the necessary libraries and define two mandatory functions:
    \begin{itemize}
        \item \texttt{add\_arguments}, which adds specific arguments for the new method in \texttt{argparse.ArgumentParser}.
        \item \texttt{run}, which executes the feature selection method and saves the reduced \texttt{dataset} at the end of execution.
    \end{itemize}
\end{enumerate}

The framework supports a diversified methodology, offering flexibility for application across different domains and datasets. 
Additionally, \mhframework provides customization features, allowing users to adapt feature selection strategies according to specific domain requirements.

Due to its simple and independent structural design, \mhframework offers intrinsic scalability, as methods are independent, and there is no limit to the number of them that can be incorporated. 
In terms of performance, the framework supports parallel execution of methods, incorporates error handling, and includes logging mechanisms, ensuring reliability and facilitating debugging during execution.

Additional details about datasets, framework implementation, installation instructions, execution environments, parameters, and other technical information are available in the \mhframework GitHub repository\footnote{\scriptsize \url{https://github.com/SBSegSF24/MH-FSF}}.

\textcolor{black}{Although the main focus of this study is Android malware detection, the MH-FSF framework is designed with sufficient modularity and flexibility to be applicable to a variety of domains, including network traffic analysis, biomedical data interpretation, and fraud detection. Its architecture allows for the replacement of datasets and the tuning of classifiers, enabling its generalization to a variety of supervised learning scenarios.}

\section{Experimentation}
\label{sec_experimentation}

To evaluate and compare the 17 feature selection methods integrated into the MH-FSF framework, we curated and utilized 10 publicly available datasets. Table \ref{tab_datasets} details these datasets, specifying the number and types of features—API calls (A), permissions (P), intents (I), and Op Codes (O)—as well as the count of benign and malicious samples. As evident, the datasets exhibit significant heterogeneity, with substantial variations in sample sizes, feature types, and feature counts.

\begingroup
\setlength{\tabcolsep}{3pt}
\begin{table}[htpb]
\caption{Summary of datasets used by \mhframework framework.}
\centering
\begin{tabular}{c||c|c||c|c||c}
\multicolumn{1}{c||}{\multirow{2}{*}{\textbf{Dataset}}} &
\multicolumn{2}{c||}{\textbf{Features}} &
\multicolumn{2}{c||}{\textbf{Complete}} & \textbf{Balanced}\\
\cline{2-6}
 & \textbf{\#} & \textbf{Types} & \textbf{M.} & \textbf{B.} & \textbf{\shortstack{Per\\Class}} \\
\hline \hline
Adroit & 166 & P & 3418 & 8058 & 3418\\
\hline
AndroCrawl & 81 & \longText{A(24)\\I(8)\\P(49)} & 10170 & 86562 & 10170\\
\hline
\longText{Android\\Permissions} & 151 & P & 17787 & 9077 & 9077\\
\hline
\longText{DefenseDroid\\PI} & 2938 & \longText{P(1490)\\I(1448)} & 6000 & 5975 & 5975\\
\hline
\longText{DefenseDroid\\A (C, D, K)} & \longText{4275,\\6003,\\6003} & A & 5254 & 5222 & 5222\\
\hline
Drebin-215 & 215 & \longText{A(73)\\P(113)\\O(6)\\I(23)} & 5560 & 9476 & 5560\\
\hline
KronoDroid R. & 246 & \longText{P(146)\\A(100)} & 41382 & 36755 & 36755\\
\hline
KronoDroid E. & 268 & \longText{P(145)\\A(123)} & 28745 & 35246 & 28745\\
\end{tabular}
\label{tab_datasets}
\end{table}
\endgroup

For the DefenseDroid dataset, we utilized four distinct variants. The first variant, DefenseDroid PI, comprises permissions (P) and intents (I). The second variant consists of three variations of API calls (A), denoted as DefenseDroid A (C, D, and K), where C, D, and K represent the application of closeness (C), degree (D), and Katz (K) techniques for data normalization to generate tabular and binary datasets.

To evaluate the datasets resulting from feature selection (reduced datasets), we employed three classifiers: KNN (clustering-based), RF (tree-based), and SVM (margin-based). For implementation and experimentation, we used the default configuration of the \texttt{scikit-learn} library, version 1.5.2 \cite{scikit-learn}.

Beyond traditional evaluation metrics (accuracy, precision, recall, and F1 score), we also incorporated the Matthews Correlation Coefficient (MCC), which quantifies the correlation between predicted and actual classes. MCC achieves high values only when the classifier performs optimally across all four cells of the confusion matrix \cite{cao2020mcc}.

Given the inherent imbalance typically observed in our datasets, we opted for stratified cross-validation, which maintains the original class proportions during evaluation. Cross-validation is commonly performed with \(K = 5\) or \(K = 10\), as these values offer test error rate estimates with controlled bias and variance \cite{james2013introduction}. We adopted the default \(K = 5\) configuration of the \textit{scikit-learn} library for our experiments.

To ensure reproducibility, we conducted all experiments on a machine equipped with an Intel(R) Xeon(R) CPU E5-4617 processor at 2.90GHz, 64GB of RAM, and 800GB of storage, running Linux Ubuntu 22.04 LTS. Additional details regarding the experimental environment are available in the MH-FSF GitHub repository.

\section{Results}
\label{sec_results}

The Table~\ref{tab_results_summary} presents the top-performing feature selection methods, ranked in descending order based on recall and F1 score metrics. These results were derived by aggregating the performance of three classifiers (KNN, RF, and SVM) across all datasets, encompassing both balanced and imbalanced scenarios.

\begingroup
\setlength{\tabcolsep}{5pt}
\begin{table}[ht!]
\caption{F1 scores and recall of methods on datasets.}
\centering
\begin{tabular}{c||c|c||c|c}
\multicolumn{1}{c||}{\multirow{2}{*}{\textbf{Method}}} & \multicolumn{2}{c||}{\textbf{Balanced}} & \multicolumn{2}{c} {\textbf{Complete}}\\ \cline{2-5}
 & \textbf{F1} & \textbf{Recall} &  \textbf{F1} & \textbf{Recall} \\ \hline \hline
ABC         & 0.85 & 0.84 & 0.72 & 0.73 \\ \hline
ANOVA       & \textbf{0.90} & 0.88 & 0.90 & 0.90 \\ \hline
Chi-Square  & \textbf{0.90} & 0.88 & 0.90 & 0.90 \\ \hline
IG          & \textbf{0.90} & 0.88 & 0.90 & 0.90 \\ \hline
JOWMDroid   & \textbf{0.90} & \textbf{0.89} & 0.65 & 0.64 \\ \hline
LASSO       & \textbf{0.90} & \textbf{0.89} & \textbf{0.91} & \textbf{0.91} \\ \hline
LR          & 0.87 & 0.86 & 0.87 & 0.87 \\ \hline
MAD         & \textbf{0.90} & 0.88 & 0.90 & 0.90 \\ \hline
MT          & 0.77 & 0.76 & 0.79 & 0.78 \\ \hline
PCA         & 0.67 & 0.63 & 0.67 & 0.66 \\ \hline
PCC         & \textbf{0.90} & 0.88 & 0.90 & 0.90 \\ \hline
RFE         & \textbf{0.90} & \textbf{0.89} & 0.90 & 0.90 \\ \hline
ReliefF     & 0.71 & 0.68 & 0.63 & 0.64 \\ \hline
RFG         & 0.72 & 0.70 & 0.87 & 0.87 \\ \hline
SemiDroid   & \textbf{0.90} & \textbf{0.89} & 0.90 & 0.90 \\ \hline
SigAPI      & 0.87 & 0.86 & 0.90 & 0.90 \\ \hline
SigPID      & 0.71 & 0.66 & 0.65 & 0.67 \\
\end{tabular}
\label{tab_results_summary}
\end{table}
\endgroup

\begin{figure}[ht!]
    \centering
    \begin{subfigure}[t]{0.49\textwidth}
        \centering
        \includegraphics[width=\textwidth]{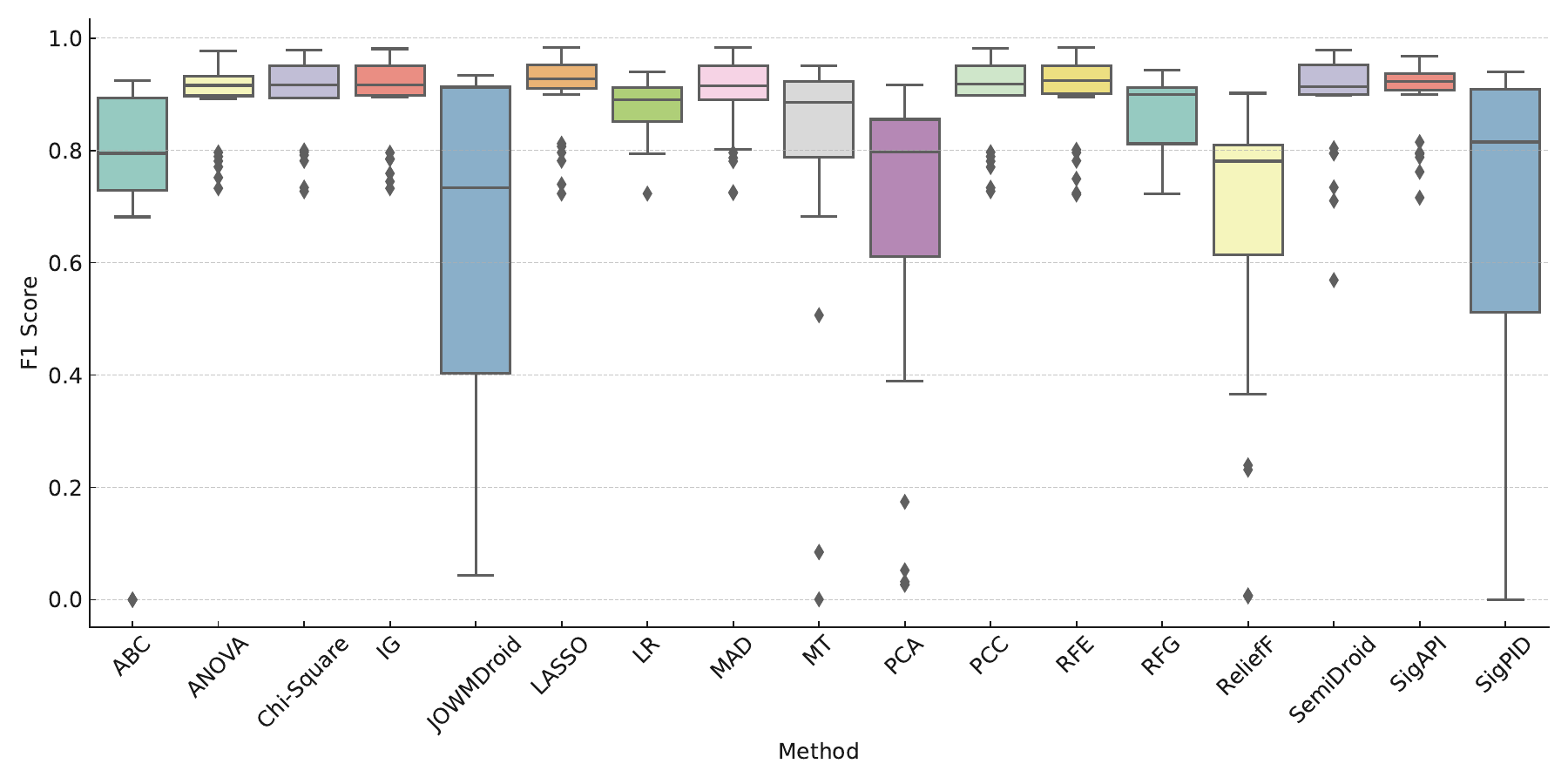}
        \caption{Complete datasets.}
        \label{fig:f1_scores}
    \end{subfigure}
    \hfill
    \begin{subfigure}[t]{0.49\textwidth}
        \centering
        \includegraphics[width=\textwidth]{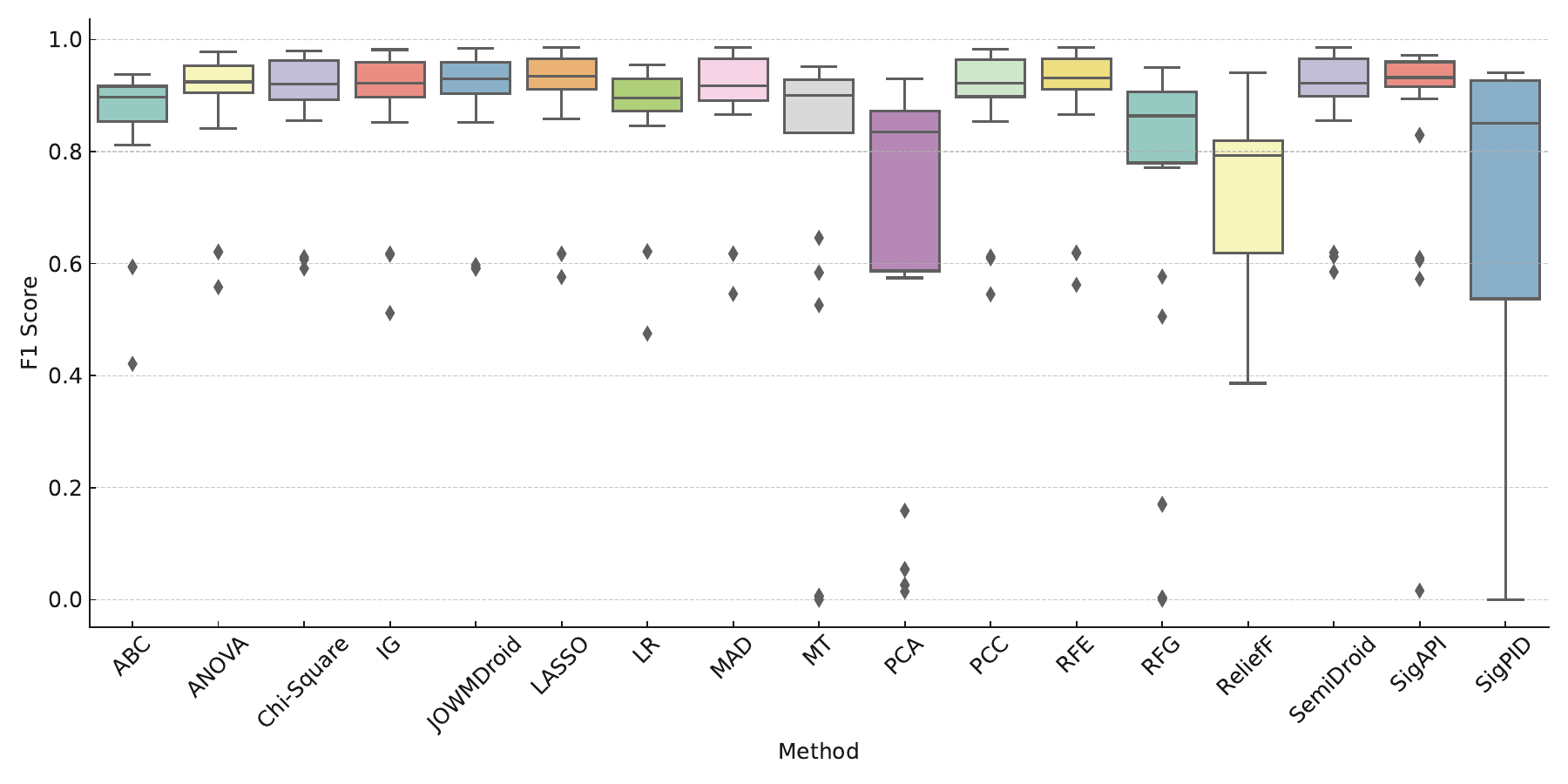}
        \caption{Balanced datasets.}
        \label{fig:recall_scores}
    \end{subfigure}
    \caption{F1 score distribuition by method.}
    \label{fig:fs_comparison}
\end{figure}

The analysis of results, complemented by the boxplots in Figure~\ref{fig:fs_comparison}, provides clear evidence of the varying effectiveness of feature selection methods across both balanced and complete datasets.

\textbf{LASSO} and \textbf{RFE} emerge as top-performing methods, consistently achieving F1 scores and recall values above 0.9. Their boxplots exhibit low variability, with compact interquartile ranges and an absence of significant outliers. This visual consistency reinforces their reliability and robustness, suggesting a strong generalization ability even across datasets with different class distributions. The effectiveness of these techniques—based respectively on regularization and wrapper strategies—lies in their ability to retain the most relevant features with high precision and recall.

\textbf{SigAPI} also stands out as a competitive and reliable method. While not always outperforming LASSO or RFE, it exhibits consistently high F1 scores across datasets, with minimal performance degradation in both balanced and imbalanced conditions. Its semantic understanding of API usage contributes to more stable selection outcomes, particularly in real-world malware detection contexts.

In contrast, methods such as \textbf{PCA}, \textbf{ReliefF}, and \textbf{SigPID} perform substantially worse, particularly on complete datasets. Their corresponding boxplots show broader interquartile ranges and noticeably lower medians, indicating higher sensitivity to data imbalance and distributional shifts. For instance, PCA’s reliance on variance as a selection criterion often leads to exclusion of highly discriminative features, especially when variance does not align with class boundaries —as clearly reflected in lower performance distributions of both F1 and recall. Similarly, ReliefF and SigPID exhibit significant variability, suggesting that their selection heuristics are less robust when faced with heterogeneous or skewed data.

Methods like \textbf{Chi-Square}, \textbf{IG}, \textbf{ANOVA}, and \textbf{MAD} demonstrate competitive mid-range performance, especially in balanced scenarios. Although they do not surpass LASSO or RFE, their visual profiles show moderate variability and relatively high medians, suggesting that they can serve as strong baseline approaches when computational simplicity or interpretability is a priority. These methods also benefit modestly from balancing, with slightly more compact score distributions observed in balanced setting.

It is also worth highlighting the performance of domain-specific methods such as \textbf{SemiDroid}, \textbf{SigAPI}, and \textbf{JOWMDroid}. Some, like SemiDroid and SigAPI, achieve results comparable to classical techniques, while JOWMDroid underperforms, as evidenced by more dispersed boxplots and occasional outliers. This inconsistency, clearly observable in visualizations, supports the notion that many domain-specific methods are not consistently benchmarked against a broad and representative set of alternatives. As a result, conclusions drawn from isolated evaluations may lack generalizability and potentially overestimate the actual performance of such methods.

Taken together, these findings reinforce importance of using robust, stable, and well-generalized feature selection techniques in malware detection pipelines. The comparative visual analysis underscores not only which methods perform best on average but also which offer consistent results across diverse data conditions, an essential criterion for real-world applicability.

\begin{figure*}[ht!]
    \centering
    \includegraphics[width=1.\textwidth]{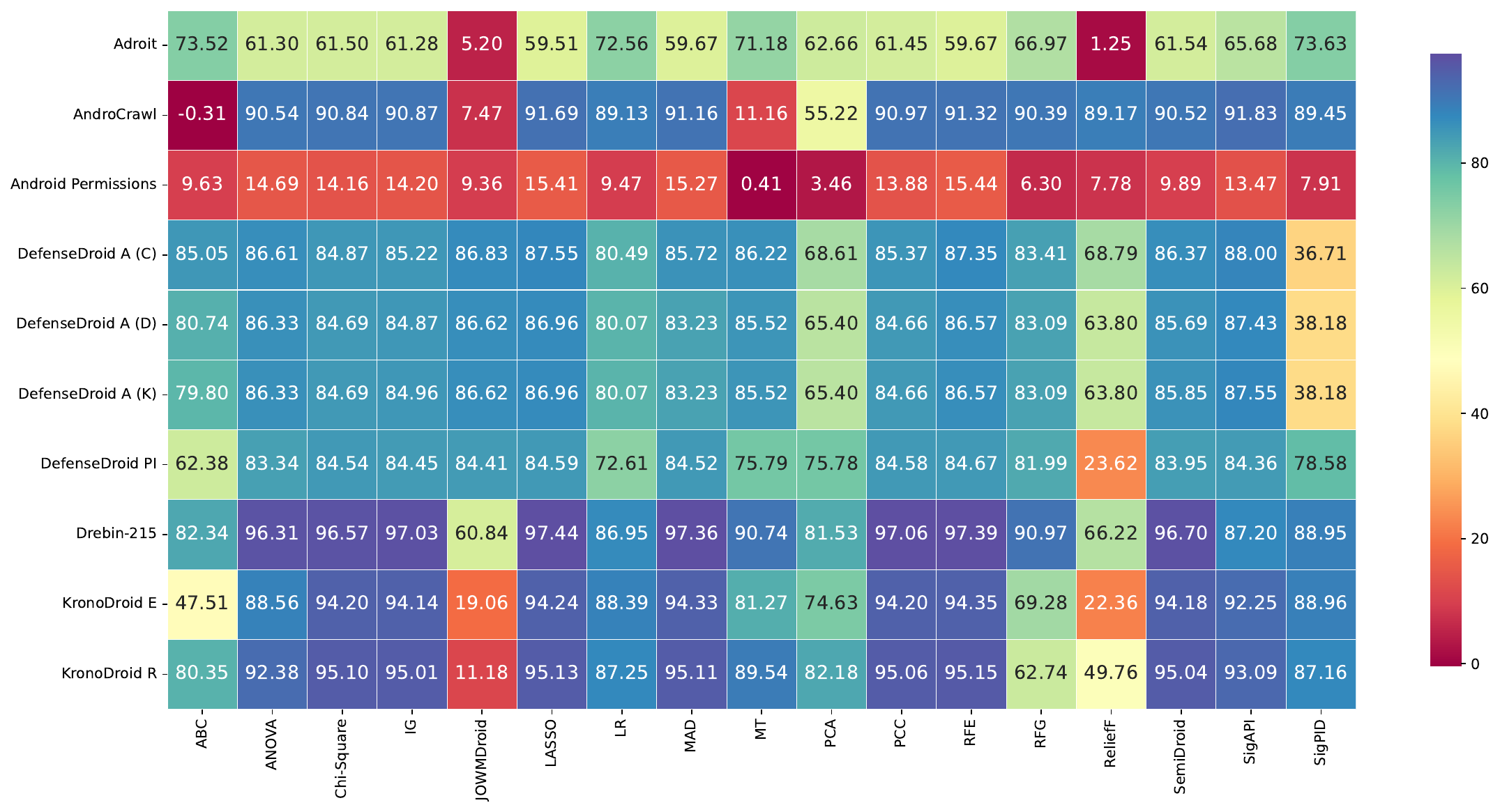}
    \caption{MCC: Complete Datasets X Feature Selection Methods.}
    \label{fig_mcc_map}
\end{figure*}

\begin{figure*}[ht!]
    \centering
    \includegraphics[width=1.\textwidth]{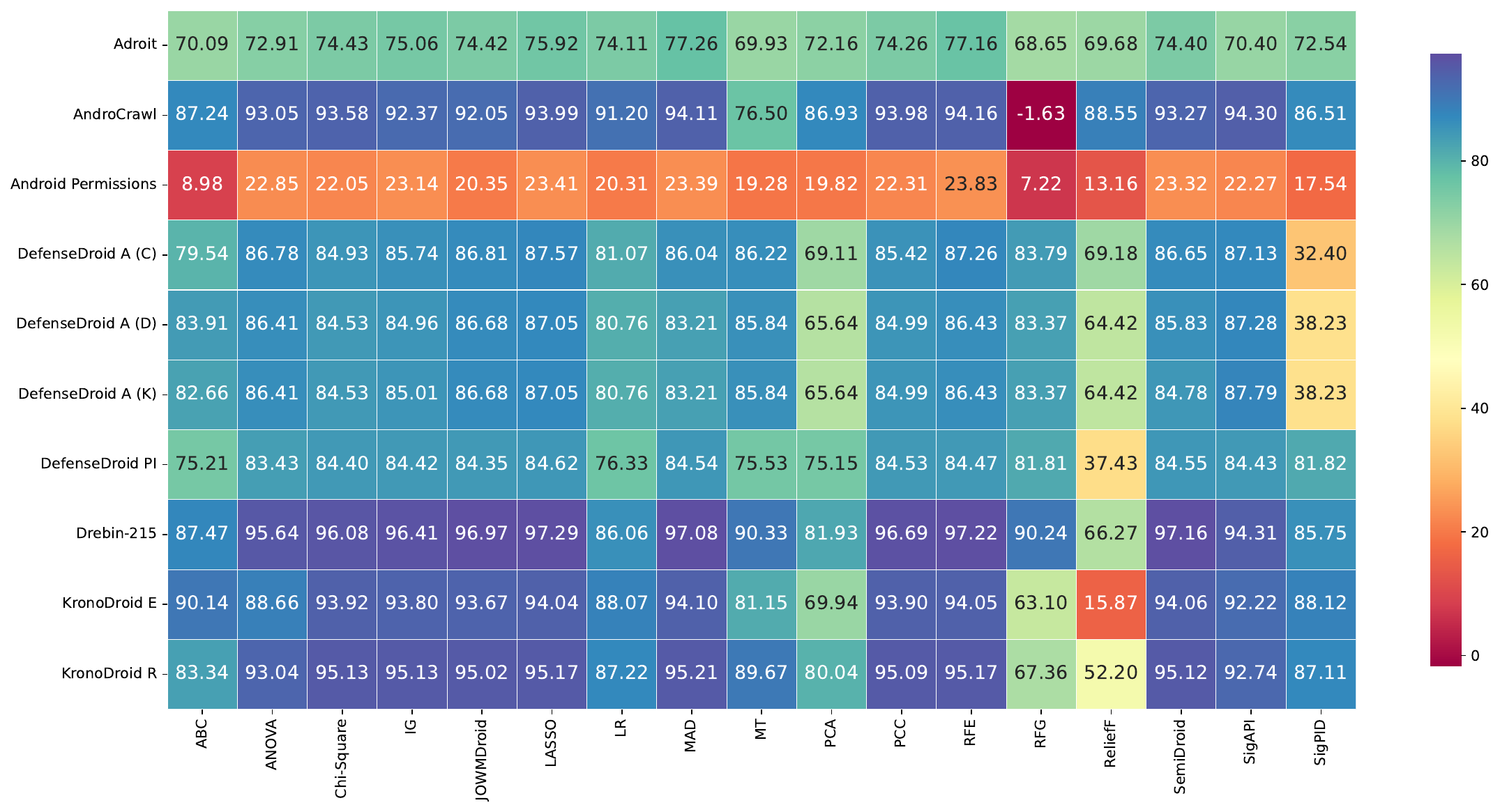}
    \caption{MCC: Balanced Datasets X Feature Selection Methods.}
    \label{fig_balanced_mcc_map}
\end{figure*}

To further elucidate the performance of these methods, we present Matthews Correlation Coefficient (MCC) heatmaps for each method across two dataset types: complete (Figure \ref{fig_mcc_map}) and balanced (Figure \ref{fig_balanced_mcc_map}). The MCC metric provides valuable insights into the effectiveness of the methods. For instance, \textbf{LASSO} and \textbf{RFE} consistently demonstrate superior performance on datasets such as KronoDroid R, KronoDroid E, and Drebin-215, across both balanced and imbalanced (complete) versions. This consistency suggests that these techniques effectively identify and select relevant features, even in presence of class imbalance.

Datasets like \textbf{Adroit}, \textbf{AndroCrawl}, and \textbf{Android Permissions} present substantial challenges for feature selection methods. Class imbalance frequently impairs the ability of these methods to effectively differentiate malicious behaviors, as demonstrated by the performance variations observed before and after dataset balancing.

\textbf{SigAPI} method achieved the highest performance on datasets comprised solely of API calls (DefenseDroid A (C), (D), and (K)), with MCC values of 88.00\%, 87.43\%, and 87.55\%, respectively. This method's specialization in handling API call features is evidenced by its consistent performance across all scenarios. Although it did not achieve the absolute highest overall performance, its narrow margin compared to LASSO and strong performance on other datasets suggest rigorous design and evaluation, resulting in robust generalization capabilities.

The variability in results across imbalanced datasets highlights the inherent complexity of real-world data analysis, where class imbalances are prevalent. Addressing this complexity necessitates appropriate data preprocessing strategies or the adoption of feature selection methods specifically designed to handle such asymmetries, thereby enhancing prediction accuracy and reliability.

Our findings highlight that proper data balancing is critical to maximizing the effectiveness of feature selection methods, as it ensures that the most relevant features are identified consistently, regardless of class distribution. 

Among evaluated techniques, \textbf{LASSO}, \textbf{RFE}, and \textbf{SigAPI} emerge as most reliable, consistently achieving high average performance with low variability across both F1 score and recall metrics. These methods exhibit an effective trade-off between precision and recall, supporting efficient and dependable malware detection. In contrast, approaches such as \textbf{ReliefF} and \textbf{JOWMDroid} reveal notable limitations, highlighting the need for further refinement or development of alternative strategies to ensure accurate classification. This analysis reinforces the importance of selecting robust and stable detection techniques to enhance overall reliability of malware detection systems, thereby contributing to more secure computing environments.

Furthermore, these results suggest a potential bias in design of methods like JOWMDroid, which were evaluated on a single proprietary dataset (unavailable for replication) and compared to only two classical methods, often in a manner that raises concerns regarding the fairness of comparisons.

\textcolor{black}{The superiority of LASSO and RFE methods is directly linked to their ability to handle correlations between features, eliminating redundant variables and promoting more generalizable models. LASSO, in particular, tends to select only one variable from a group of highly correlated ones, which helps simplify the model without significantly compromising the represented information. On the other hand, methods such as PCA and ReliefF showed inferior performance on more imbalanced datasets, as they do not explicitly account for importance of minority classes. Methods like JOWMDroid and SigPID were also negatively affected, mainly due to their strong dependence on specific types of features — such as permissions — which limits their effectiveness in datasets dominated by API calls.}

\section{Discussion}

First, it is crucial to emphasize that evaluating feature selection methods using a representative and substantial number of datasets (e.g., ten) is essential for developing effective models for Android malware classification. Such evaluations provide valuable insights into the strengths and weaknesses of various methods, guiding future research directions.

We now address the key research questions that guided our study. The answers and findings aim to provide valuable insights for researchers advancing feature selection methods in the context of Android malware detection.

\noindent \emph{Are domain-specific feature selection methods more efficient than classical ones in detecting Android malware?}

In general, certain classical feature selection methods have demonstrated greater efficiency compared to domain-specific methods. However, both types of methods possess distinct advantages and disadvantages. While most feature selection methods can identify effective feature subsets, they may struggle with complex datasets or those with a high number of features. This can lead to suboptimal performance when applied to datasets using features outside their design scope. For example, SigPID, which specializes in permissions, encounters difficulties when applied to API-based datasets.

Proper feature selection significantly enhances the performance of classification models. Therefore, it is critical to carefully consider dataset characteristics before selecting a method for Android malware detection. Furthermore, combining different algorithms and techniques may yield optimal results.

\noindent \emph{Which is the best feature selection method for Android malware detection?}

Based on the Matthews Correlation Coefficient (MCC) results presented in the heatmap in Figure \ref{fig_mcc_map}, LASSO demonstrated strong performance across nearly all datasets, indicating its potential as a robust feature selection method for detecting malicious samples. LASSO exhibited consistent MCC values, making it a reliable candidate for feature selection in this domain.

However, LASSO exhibited lower performance on the Adroit dataset. We attribute this to LASSO's tendency to select a single variable from highly correlated variable groups, potentially overlooking others. Given the high correlation among features in the Adroit dataset, LASSO may fail to select all relevant features, leading to information loss.

Datasets such as Adroit, AndroCrawl, and Android Permissions present challenges for feature selection methods. Methods that excelled on balanced versions of these datasets (e.g., ABC, JOWMDroid, PCA) demonstrated superior performance, as evidenced by higher MCC values. This suggests that class imbalance in imbalanced data can compromise the ability of feature selection methods to effectively differentiate malicious behaviors, as demonstrated by the improved results after balancing.

The variability in results across imbalanced datasets underscores the inherent complexity of real-world data analysis, where class imbalances are prevalent. Addressing this complexity necessitates appropriate data preprocessing strategies or the adoption of feature selection methods specifically designed to handle such asymmetries, thereby enhancing prediction accuracy and reliability.

Our results demonstrate that proper data balancing is crucial for maximizing the efficacy of feature selection methods, ensuring the identification of the most relevant features regardless of class distribution.

Finally, we have made the complete results—for all 17 methods, 10 datasets, and 3 models—available in MH-FSF framework repository on GitHub \cite{rocha2024mhfsf}. This enables readers and researchers to conduct in-depth analyses and discussions regarding specific methods or datasets.

\section{Final Considerations and Future Directions}
\label{sec_final}

In this work, we introduced the MH-FSF framework to address the limitations of current feature selection research, specifically the lack of robust benchmarking and reproducibility due to limited comparisons and reliance on proprietary datasets. MH-FSF provides a comprehensive, modular, and extensible platform for the reproduction, implementation, and evaluation of diverse feature selection methods, including 11 classical and 6 domain-specific techniques, across 10 publicly available Android malware datasets.

Our findings underscore the critical need for unified platforms like MH-FSF to facilitate direct and comprehensive comparisons of feature selection methods. By demonstrating performance variations across balanced and imbalanced datasets, we highlighted the importance of data preprocessing and selection criteria that account for class asymmetries. The availability of MH-FSF on GitHub \cite{rocha2024mhfsf}, including all implementations, experiment automation scripts, datasets, and detailed results, ensures transparency and reproducibility, fostering methodological consistency and rigor in future research.

Future research directions may include:

\begin{enumerate}
    \item Expanding the MH-FSF framework with additional feature selection methods to broaden its scope and applicability.
    \item Evaluating feature selection methods on contemporary and updated Android malware datasets to ensure relevance to current threats.
    \item Applying explainable AI (XAI) techniques to elucidate the behavior of feature selection methods and enhance the interpretability of results.
    \item Assessing the performance of feature selection methods in real-time data collection environments to evaluate their practical applicability.
    \item Investigating the impact of adversarial attacks on feature selection to understand and mitigate potential vulnerabilities.
\end{enumerate}

By pursuing these directions, we aim to further broaden the existing literature and pave the way for new research advancements in feature selection, particularly within the context of Android malware detection.

\section*{Declarations}

\subsection*{Author Contributions}

Vanderson Rocha and Diego Kreutz contributed to the conception of this study.  
Vanderson Rocha and Hendrio Bragança performed the experiments.  
Diego Kreutz, Hendrio Bragança, and Eduardo Feitosa acted as reviewers, providing revisions and supervision.  
Vanderson Rocha is the main contributor and writer of this manuscript.  
All authors read and approved the final manuscript.

\subsection*{Competing Interests}

The authors declare that they have no competing interests.

\subsection*{Funding}

This research was partially funded, as stipulated in Articles 21 and 22 of Decree No. 10.521/2020, under Federal Law No. 8.387/1991, through agreement No. 003/2021, signed between ICOMP/UFAM, Flextronics da Amazônia Ltda., and Motorola Mobility Comércio de Produtos Eletrônicos Ltda.  
This work was also supported by the Coordination for the Improvement of Higher Education Personnel – Brazil (CAPES) – Financing Code 001, and partially supported by the Amazonas State Research Support Foundation (FAPEAM) through the POSGRAD project 2024/2025.  
This research was also partially funded by FAPERGS through grant agreements 24/2551-0001368-7 and 24/2551-0000726-1.

\subsection*{Availability of Materials}

The source code and datasets analyzed in this study are publicly available in the MH-FSF GitHub repository:  
\url{https://github.com/SBSegSF24/MH-FSF}

\bibliographystyle{IEEEtran}
\bibliography{refs}

\end{document}